\pdfoutput=1

\documentclass[11pt]{article}

\usepackage[final]{acl}

\usepackage{times}
\usepackage{enumitem}
\usepackage{float}  
\usepackage{latexsym}
\usepackage{amsthm}
\usepackage{xcolor}
\usepackage{booktabs}
\definecolor{PineGreen}{rgb}{0.0, 0.47, 0.32} 
\definecolor{SpringGreen}{rgb}{0.0, 1.0, 0.5}
\definecolor{OliveGreen}{rgb}{0.5, 0.5, 0.0} 
\definecolor{Violet}{rgb}{0.933, 0.509, 0.933} 
\definecolor{Plum}{rgb}{0.866, 0.627, 0.867} 
\definecolor{DeepPlum}{rgb}{0.412, 0.215, 0.412} 
\newcounter{example} 

\usepackage{enumitem}
\renewcommand
\newcounter{example} 

\newlist{examplelist}{enumerate}{1} 
\setlist[examplelist]{label=\arabic{example}.,ref=\arabic{example}}

\usepackage[T1]{fontenc}

\usepackage[utf8]{inputenc}

\usepackage{microtype}

\usepackage{inconsolata}

\usepackage{graphicx}

\title{\textit{SPACER}: A Parallel Dataset of \textit{S}peech \textit{P}roduction \textit{A}nd \textit{C}omprehension of \textit{E}rror \textit{R}epairs}

 \author{Shiva Upadhye* \hspace{1em}  Jiaxuan Li* \hspace{1em}  Richard Futrell \\
         $\texttt{\{shiva.upadhye, jiaxuan.li, rfutrell\}@uci.edu}$ \\
         Department of Language Science \\
         University of California, Irvine \\
         \small \emph{*equal contribution}}
\usepackage{hyperref}
\graphicspath{ {figures/} }
\begin{document}
\maketitle
\begin{abstract}
Speech errors are a natural part of communication, yet they rarely lead to complete communicative failure because both speakers and comprehenders can detect and correct errors. Although prior research has examined error monitoring and correction in production and comprehension separately, integrated investigation of both systems has been impeded by the scarcity of parallel data. In this study, we present \emph{SPACER}, a parallel dataset that captures how naturalistic speech errors are corrected by both speakers and comprehenders. We focus on single-word substitution errors extracted from the Switchboard corpus, accompanied by speaker's self-repairs and comprehenders' responses from an offline text-editing experiment. Our exploratory analysis suggests asymmetries in error correction strategies: speakers are more likely to repair errors that introduce greater semantic and phonemic deviations, whereas comprehenders tend to correct errors that are phonemically similar to more plausible alternatives or do not fit into prior contexts. Our dataset~\footnote{The dataset and code are available at: \url{https://github.com/goldengua/SPACER-CMCL}}  
enables future research on integrated approaches toward studying language production and comprehension. \end{abstract} 

\section{Introduction}
Production errors are common in naturalistic speech; however, they rarely lead to a complete breakdown in communication, as interlocutors are able to monitor, detect, and repair errors in real-time. For this reason, characterizing the process of error correction has remained a shared goal of both language comprehension and production research. 

Comprehenders process errors by integrating perceived linguistic input with prior context and expectations, and might arrive at an interpretation different from the literal meaning of the linguistic input \citep{ferreira_good-enough_2002,ferreira_misinterpretation_2000,dempsey2023nonce,bader2018misinterpretation,levy_eye_2009,gibson_rational_2013}. One possible mechanism is that they perform rational inference over the perceived errors \citep{levy_eye_2009,levy2008noisy,gibson_rational_2013,futrell_lossycontext_2020,ryskin2018comprehenders,poppels2016structure,zhang2023noisy}. When the perceived form is incongruent with prior context or similar to a more plausible alternative, comprehenders might override the literal input and reconstruct an alternative interpretation. 

In language production research, speech errors have played a crucial role in shaping our understanding of the cognitive machinery of production, including the role of online control \citep{fromkin1973speech,levelt1983monitoring,dell1986spreading}. Numerous studies analyzing the temporal and distributional properties of speech errors have found evidence of a two-stage monitoring and correction process, which operates first on internal representations and then on the articulated linguistic signal \citep{levelt1983monitoring,blackmer1991theories,hartsuiker2001error,nooteboom2017self}. Although the mechanism of monitoring has remained a point of contention in the literature \citep{levelt1999producing,nozari2011comprehension,hickok2011sensorimotor,roelofs2020correctly,gauvin2020towards}, accounts of repair processing have posited sustained competition between activated representations and selection control as potential mechanisms of correction \citep{hartsuiker2001error,nozari2016cognitive,nooteboom2019temporal,gauvin2020towards}.


Much of our understanding of speakers and comprehenders' error correction strategies comes from research traditions that have made limited contact with each other. In particular, existing datasets or experimental paradigms focus solely on corrective behavior in the \emph{absence} of an interlocutor. However, in a communicative context, speaker choices may exhibit a balance between ease of production and communicative efficiency \citep{ferreira2000effect,jaeger2006speakers,jaeger2010redundancy,koranda2018word,goldberg2022good,futrell2023information}. For example, speakers may preemptively hyperarticulate words to improve comprehensibility \citep{aylett2004smooth,arnold2012audience,meinhardt2020speakers} or modulate the acoustic characteristics of their speech in response to listener feedback \citep{pate2015talkers,buz2016dynamically}. Comprehension can be talker-specific as well: comprehenders can tailor their expectations to the speaker \citep{ryskin2020talker} and adapt their error correction strategies accordingly \cite{futrell2017l2,brehm2019speaker}.


In this study, we present a parallel dataset of Speech Production and Comprehension Error Repairs (\textit{SPACER}) that captures how naturalistic speech errors are corrected by speakers and comprehenders. First, we compiled a corpus of naturalistic utterances with single-word substitution errors and repairs, as well as utterances that are not corrected by speakers. These utterances were then presented to comprehenders in a web-based text editing experiment, where each case is annotated by four to six comprehenders. Our dataset contains 1056 instances of naturalistic speaker's utterances as well 5808 comprehenders' responses to speaker's initial utterances. We also provide an exploratory analysis on how well comprehender's error correction behavior can be predicted by lexical properties of speaker's error and repairs, and vice versa. Our results suggest that asymmetries between error correction behaviors by comprehenders and speakers might be related to interaction between the two modes. The dataset offers resources to build experimental and computational work that bridges comprehension and production and informs how interaction affects strategic cue weighting in error monitoring.

\begin{figure*}[!thb]
    \centering
    \includegraphics[width=0.9\linewidth]{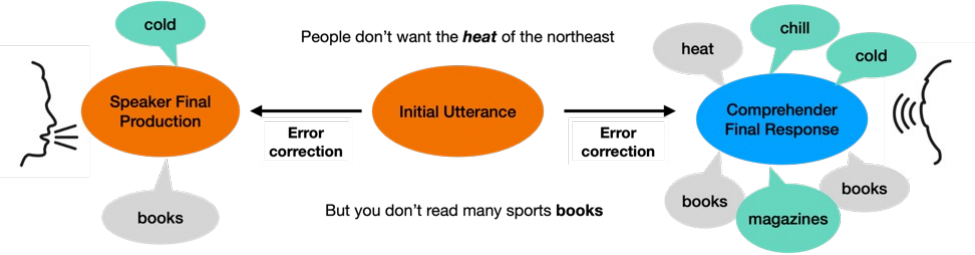}
    \caption{An illustration of the dataset design. Suppose a speaker produces an initial utterance. Both the speaker and the comprehender may engage in error monitoring and correction processes, resulting in speaker final production and comprehender final response being either the same as the initial utterance (represented in gray bubbles) or different from it (represented in green bubbles). Each utterance is annotated by four to six comprehenders.}
    \label{fig:overview}
\end{figure*}

\section{Dataset}
We focus on how comprehenders engage with utterances that may or may not have been corrected by a speaker in their original context. Figure \ref{fig:overview} illustrates the design of the dataset. We assume that both the speaker and the comprehender can perform error correction, and we remain agnostic to the mechanism that accomplishes this process. Suppose a speaker produces an initial utterance (\textit{... people don't want the heat of the northeast}), which may contain an error, or non-optimal choice of words. The speaker can monitor the initial utterance and \emph{may} correct \emph{heat} into \emph{cold}. After receiving the initial utterance, comprehenders engage in an interpretation process, where the final response might not be the same as the literal meaning of the initial utterance.

The development of this dataset involved a two-step process. First, we identified and extracted utterances with and without word substitutions and overt repairs from the Switchboard corpus \citep{godfrey1992switchboard} of spontaneous speech. Subsequently, these utterances were presented as stimuli to participants in a web-based correction experiment. Hence, for each utterance, this dataset provides a speaker's final production along with a group of comprehenders' annotations. The two stages of this process are detailed below.

\subsection{Corpus of Naturally-Occurring Utterances}
We identify and extract stimuli for the correction experiment from Switchboard NXT annotations \citep{Calhoun2010TheNS}, a subset of the Switchboard corpus that provides gold-standard disfluency annotations generated by human raters. While words in the corpus are annotated as \emph{fluent}, \emph{reparandum}, and \emph{repair}, the \emph{reparandum} label encompasses a variety of disfluencies such as filled pauses, false starts, repetitions, and substitutions. We programmatically identify substitutions using the following criteria. First, we only consider utterances with an equal number of \emph{reparandum} and \emph{repair} annotations to filter out instances where the speaker's utterance plan may have undergone structural revisions. Next, we focused on utterances where (i) the word labeled \emph{reparandum} (bolded) was immediately followed by a \textbf{non-identical}  word that was labeled \emph{repair} (underlined; see \ref{ex:fastSub}) or where (ii) the speaker repeated the \emph{reparandum} sequence almost verbatim except for a single-word change (\ref{ex:slowSub}). Finally, we eliminated instances where the \emph{reparandum} was either a filled pause, false start, repetition, or a contracted form \footnote{We eliminate instances of contractions such as \emph{I've} since they are reduced forms of multiword expressions such as \emph{I have}}. 

\begin{examplelist}
\stepcounter{example}
\item I I think that might be \textbf{talking} \underline{referring}  to uh something kind of uh alternative to the draft you know \label{ex:fastSub}
\stepcounter{example}
\item So until I see the entire quote old guard of the Soviet \textbf{military} of the Soviet \underline{government} completely roll over and disappear preferably buried I still consider them a threat \label{ex:slowSub}
\end{examplelist}


For utterances with multiple substitutions and repairs, we generate variants of the original utterance with each containing only one reparandum-repair position (see Appendix \ref{app:preprocess} for examples).

\begin{figure*}[htb!]
    \centering
\includegraphics[width=0.8\linewidth]{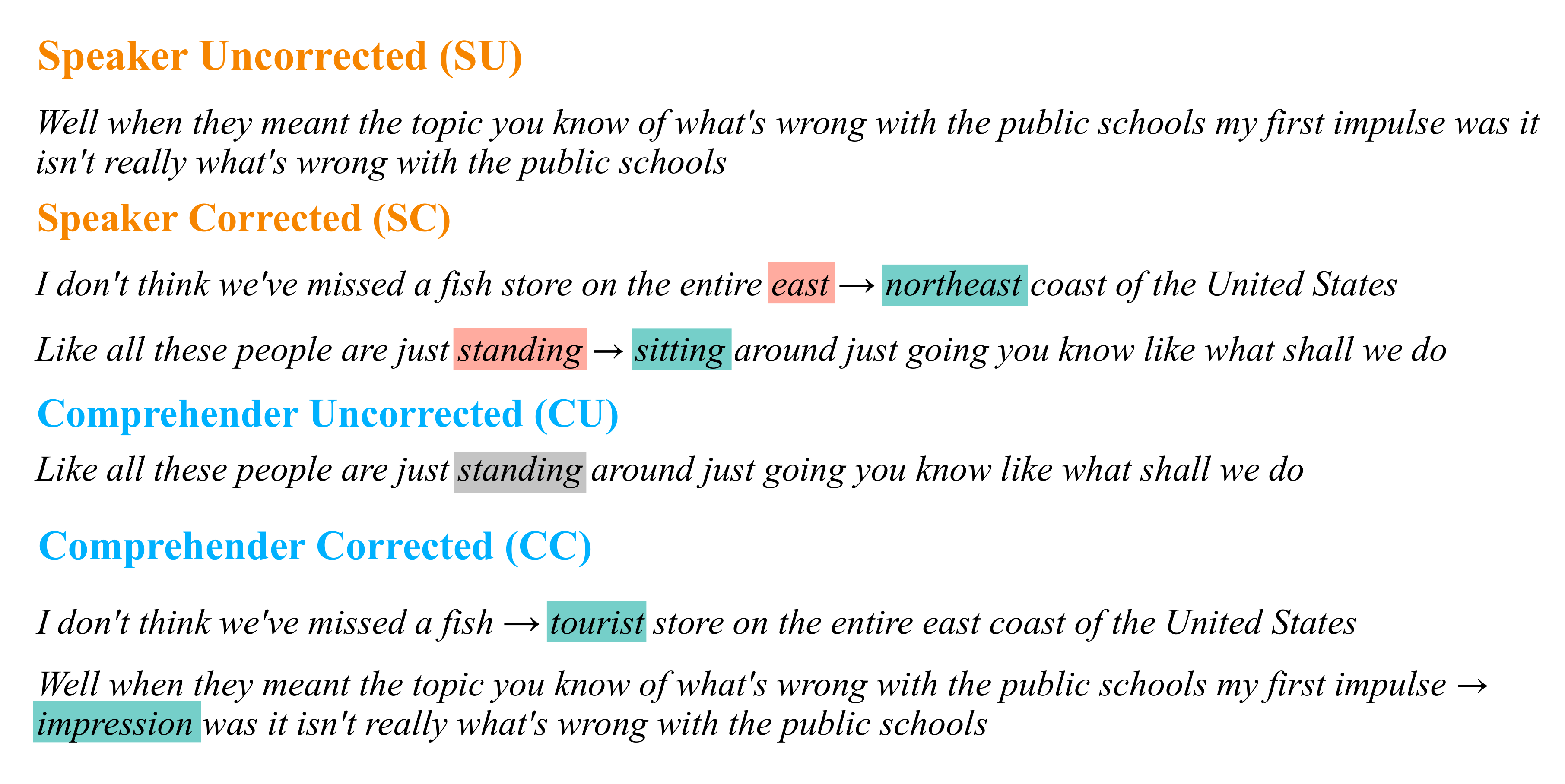}
    \caption{Examples of \emph{speaker uncorrected} (SU), \emph{speaker corrected} (SC), \emph{comprehender uncorrected} (CU), and \emph{comprehender corrected} (CC) utterances in the SPACER dataset. Words highlighted in red were initially produced a speaker and later corrected. Words highlighted in green are corrections made by either the speaker in the original context or by a participant in the comprehension experiment. A grey highlight indicates that the word was not corrected by the participant. Note that for each SU and SC utterance, there may be up to four responses, which we classify as either CC or CU responses depending on whether or not the participant made a correction in their response.}
    \label{fig:utterance-types}
\end{figure*}

Finally, the selected utterances were screened for inappropriate content by three human annotators. This process yielded a total of 576 distinct utterance frames that featured errors repaired by the speaker (henceforth, \emph{speaker corrected} or \textbf{SC} utterances). Furthermore, we also included 480 sentences from the same corpus that did not feature any overt repair made by the speaker (henceforth, \emph{speaker uncorrected} or \textbf{SU} utterances; see Fig. \ref{fig:utterance-types}).

\subsection{Correction Experiment}
We conducted human error correction experiments to understand how selected utterances are corrected during comprehension. 66 native English speakers participated in the experiment. Participants were recruited online via Prolific and compensated at \$16/hr. The experiment takes around 30 minutes. 

The stimuli consisted of 1056 initial utterances, which included both the SC (\emph{speaker corrected}) and SU (\emph{speaker uncorrected}) utterances. We distributed the 1056 selected utterances into 12 lists. Each list contains 48 SC utterances and 40 SU utterances. Each list is annotated by four to six subjects.

\begin{figure}[H]
    \centering
\includegraphics[width=0.95\linewidth]{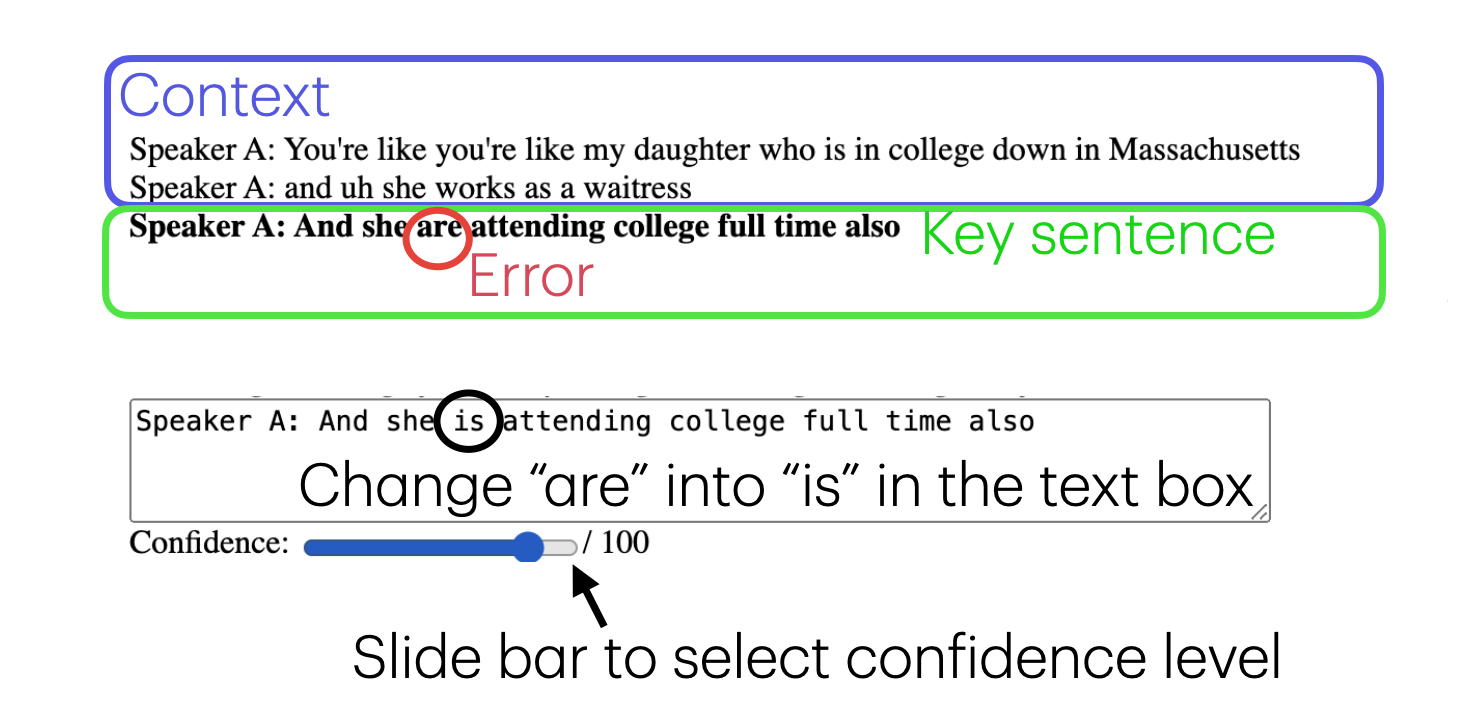}
    \caption{An illustration of comprehension experiment. A comprehender is presented with the \textit{key sentence} together with preceding \textit{context}. The comprehender is instructed to make necessary edits in the textbox and slide bar to indicate their confidence level.}
    \label{fig:exp_illustration}
\end{figure}

The subjects are presented with initial utterances along with proceeding context, and are instructed to check the quality of the last sentence from speech transcriptions and make necessary corrections by replacing the erroneous word with a more appropriate choice (Fig. \ref{fig:exp_illustration}). They also received explicit instruction to avoid insertion or deletion of word(s) (see Appendix \ref{app:experiment} for the detailed instructions provided to participants).

We exclude subjects that made less than two corrections throughout the experiment ($N = 3$), and subjects who did not move confidence bars ($N = 3$). After subject exclusion, 528 trials were further removed because they contain word insertion or deletion, resulting in 5808 responses with either one or no substitution. 

\section{Analysis}

\subsection{Descriptive summary}
As shown in Table \ref{tab:shrunk_table}, our dataset contain a total of 1056 initial utterances, where 576 initial utterances have been corrected by speakers (SC utterances). Each initial utterance is annotated by four to six comprehenders, yielding a total of 5808 responses with either one or no substitution. 34.7\% of trials were corrected by the comprehender.

\begin{table}[htb!]
    \centering
    \resizebox{\columnwidth}{!}{%
    \begin{tabular}{lrrr}
        \toprule
        & \multicolumn{2}{c}{Comprehender} & \\
        \cline{2-3}
        & Corrected & Uncorrected & Total \\
        \midrule
        Speaker Corrected & 1437 & 1731 & 3168 \\
        Speaker Uncorrected & 578 & 2062 & 2640 \\
        Total & 2015 & 3793 & 5808 \\
        \bottomrule
    \end{tabular}%
    }
    \caption{Instances of \emph{speaker corrected} and \emph{speaker uncorrected} utterances that were corrected or remained uncorrected by participants in the correction experiment}
    \label{tab:shrunk_table}
\end{table}

\begin{figure}[!htb]
    \centering
    \includegraphics[width=0.95\linewidth]{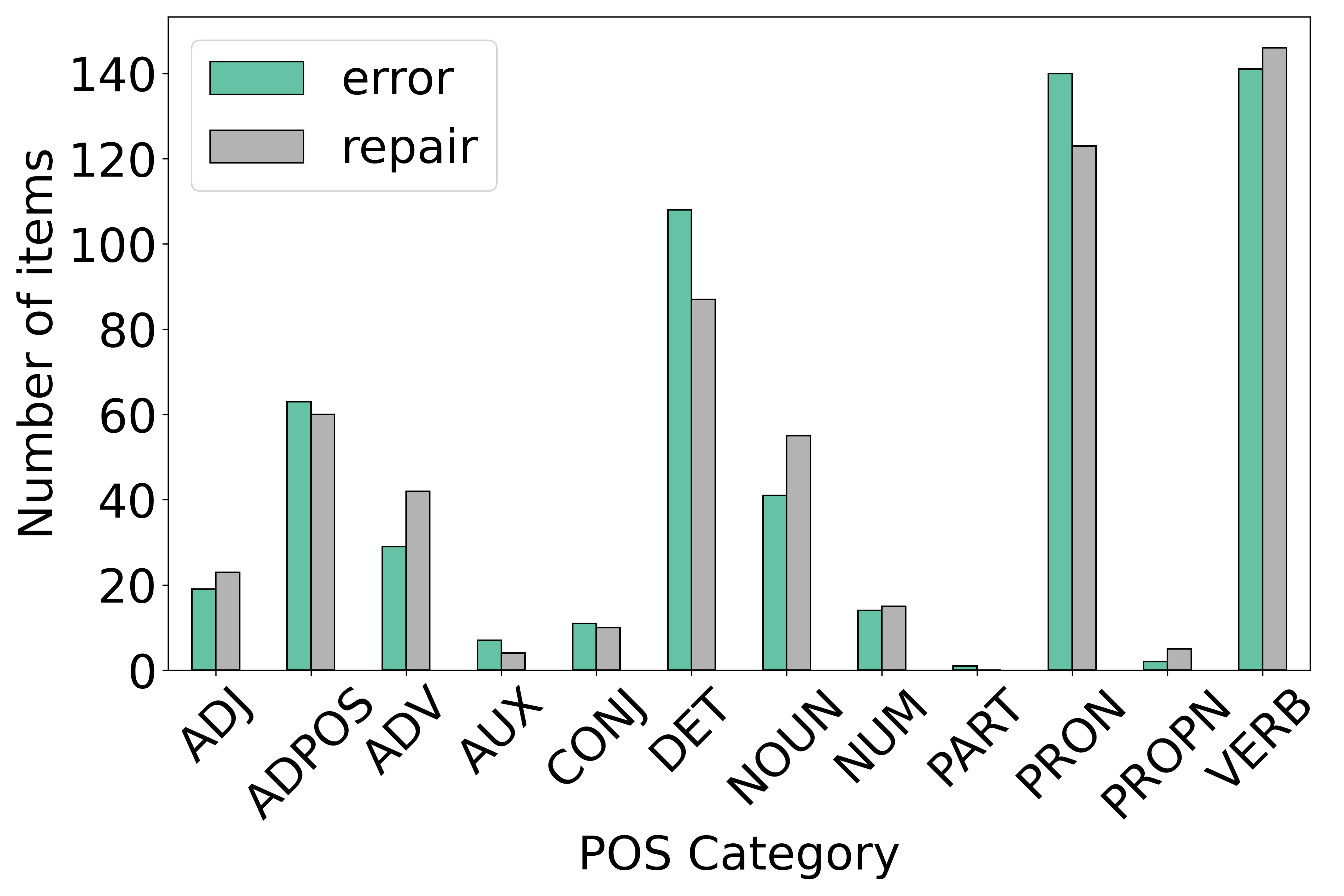}
    \caption{The POS categories of speaker-produced errors and corresponding repairs in \textit{speaker corrected} utterances.}
    \label{fig:speaker_pos}
\end{figure}

We focus on the items that are corrected by speakers (SC), and analyzed lexical properties of the critical target that has undergone correction (\textit{error}) and its corresponding corrected form (\textit{repair}).  We first analyzed the part-of-speech (POS) categories of the critical words (error and repair). Figure \ref{fig:speaker_pos} shows the distribution of part-of-speech categories for speaker initial produced errors and the corresponding repairs. The errors and repairs vary by POS category, with determiners (DET), pronouns (PRON), and verbs (VERB) exhibiting the highest frequency of both errors and repairs, whereas auxiliaries (AUX), particles (PART), and proper nouns (PROPN) show relatively fewer occurrences.

\begin{figure}[h!]
    \centering
    \includegraphics[width=0.95\linewidth]{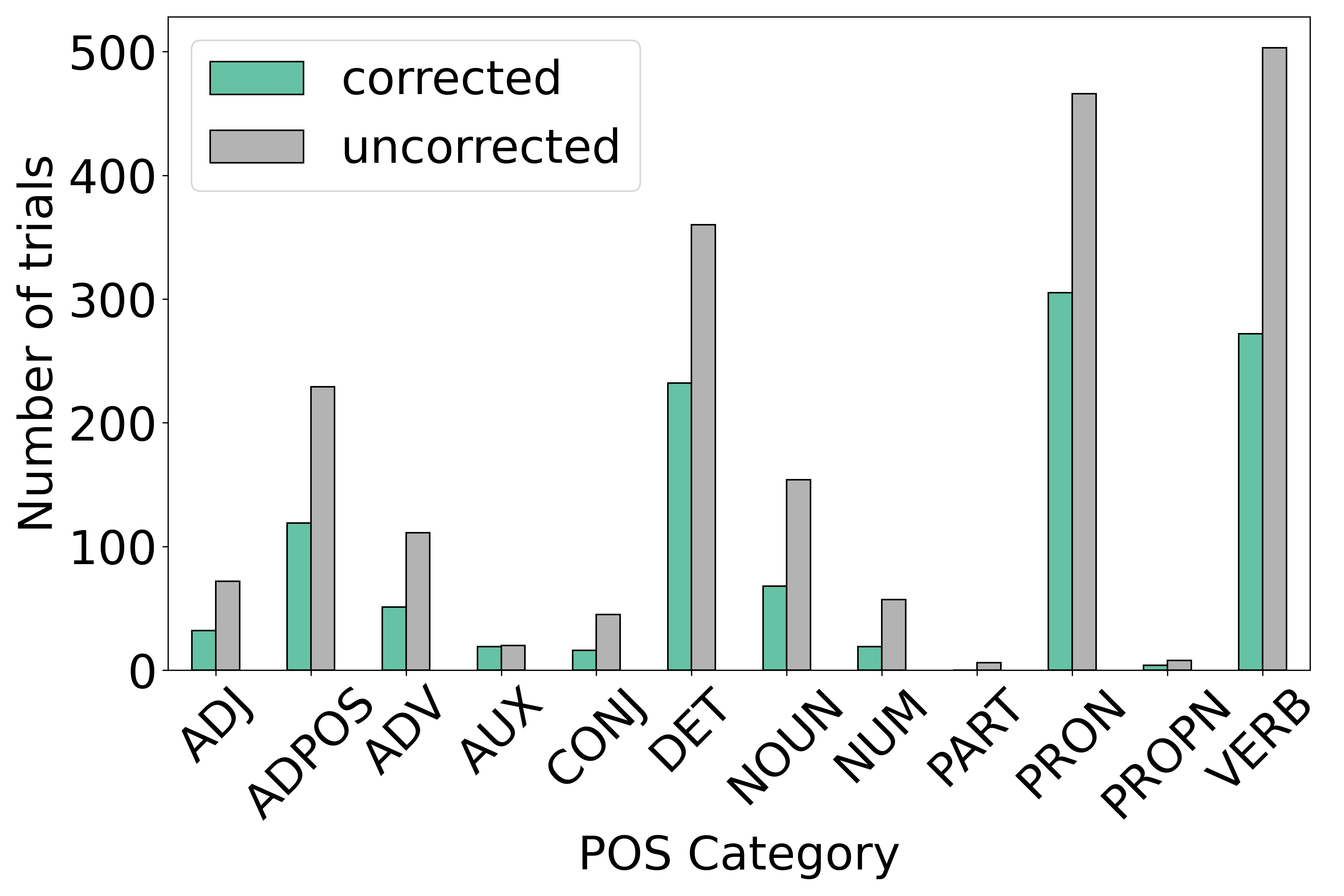}
    \caption{The number of corrected and uncorrected responses across different part-of-speech (POS) categories of presented errors in the \textit{speaker corrected} utterances.}
    \label{fig:response_pos}
\end{figure}

We further analyzed how the POS categories of speaker-produced errors would affect comprehender's error correction behavior. Figure \ref{fig:response_pos} shows the number of corrected and uncorrected responses given speaker-produced errors with different part-of-speech (POS) categories. The results suggest variation in comprehenders' tendency to correct errors depending on the POS category of the presented errors, with higher correction rates for determiners (DET), verbs (VERB), and pronouns (PRON).

\begin{figure}[htb!]
    \centering
    \includegraphics[width=0.95\linewidth]{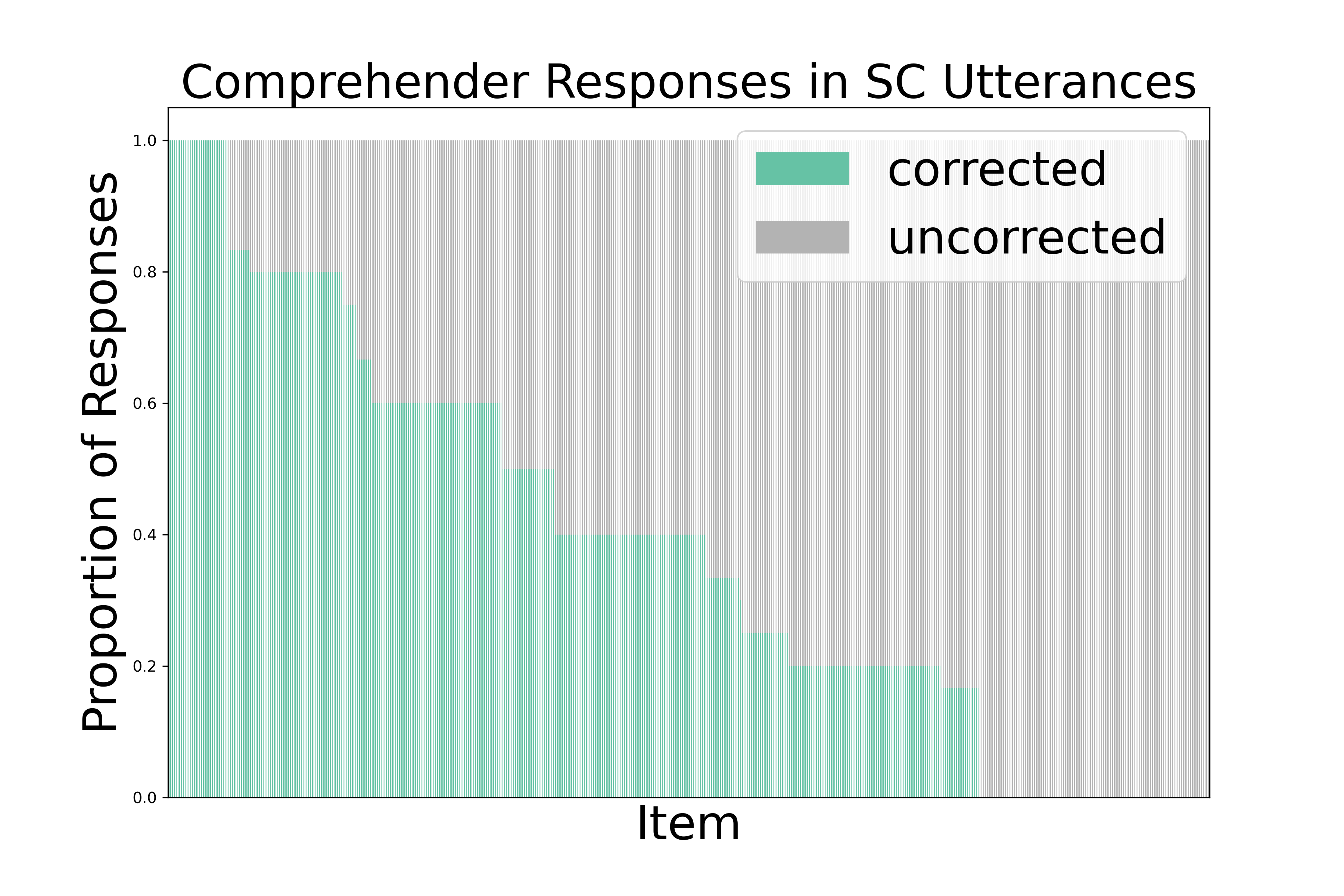}
    \includegraphics[width=0.95\linewidth]{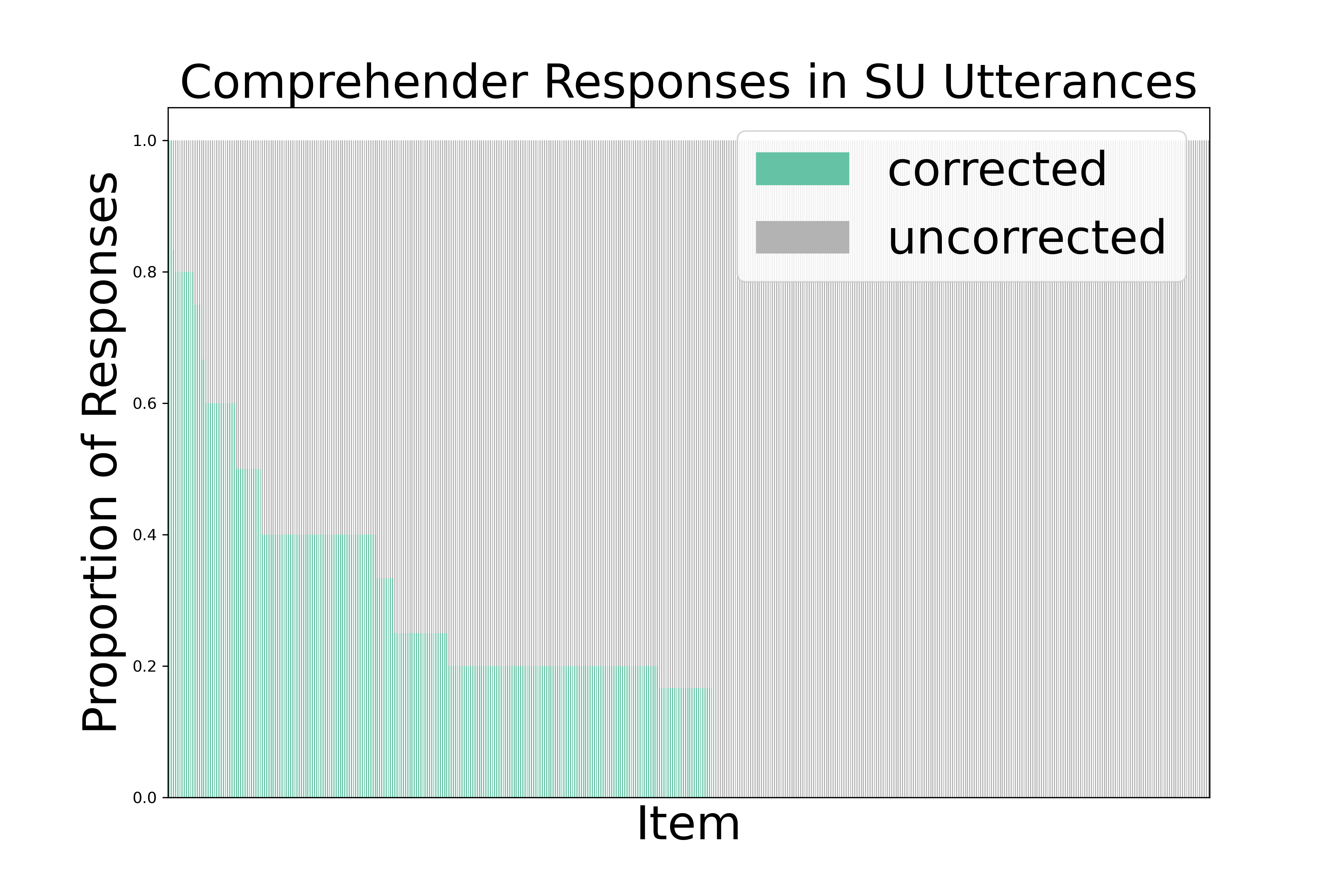}
    \caption{The proportion of corrected and uncorrected responses for each item in the \textit{speaker corrected} condition (top) and \textit{speaker uncorrected} condition (bottom). Each bar represents an individual item, with corrected responses shown in teal and uncorrected in gray.}
    \label{fig:response_condition}
\end{figure}

We examined how often comprehenders corrected errors in speaker-corrected versus speaker-uncorrected utterances. Figure \ref{fig:response_condition} presents the proportion of corrected and uncorrected responses across items in the \textit{speaker uncorrected} items (top panel) and \textit{speaker corrected} items (bottom panel). A higher proportion of items were corrected in the \textit{speaker corrected} items than in \textit{speaker uncorrected} items. While initial utterances that are corrected by speakers are also more likely to be corrected by comprehenders, there is great variation between items. 

\subsection{Inter-rater agreement}
\begin{figure}[htb!]
    \centering
    \includegraphics[width=0.95\linewidth]{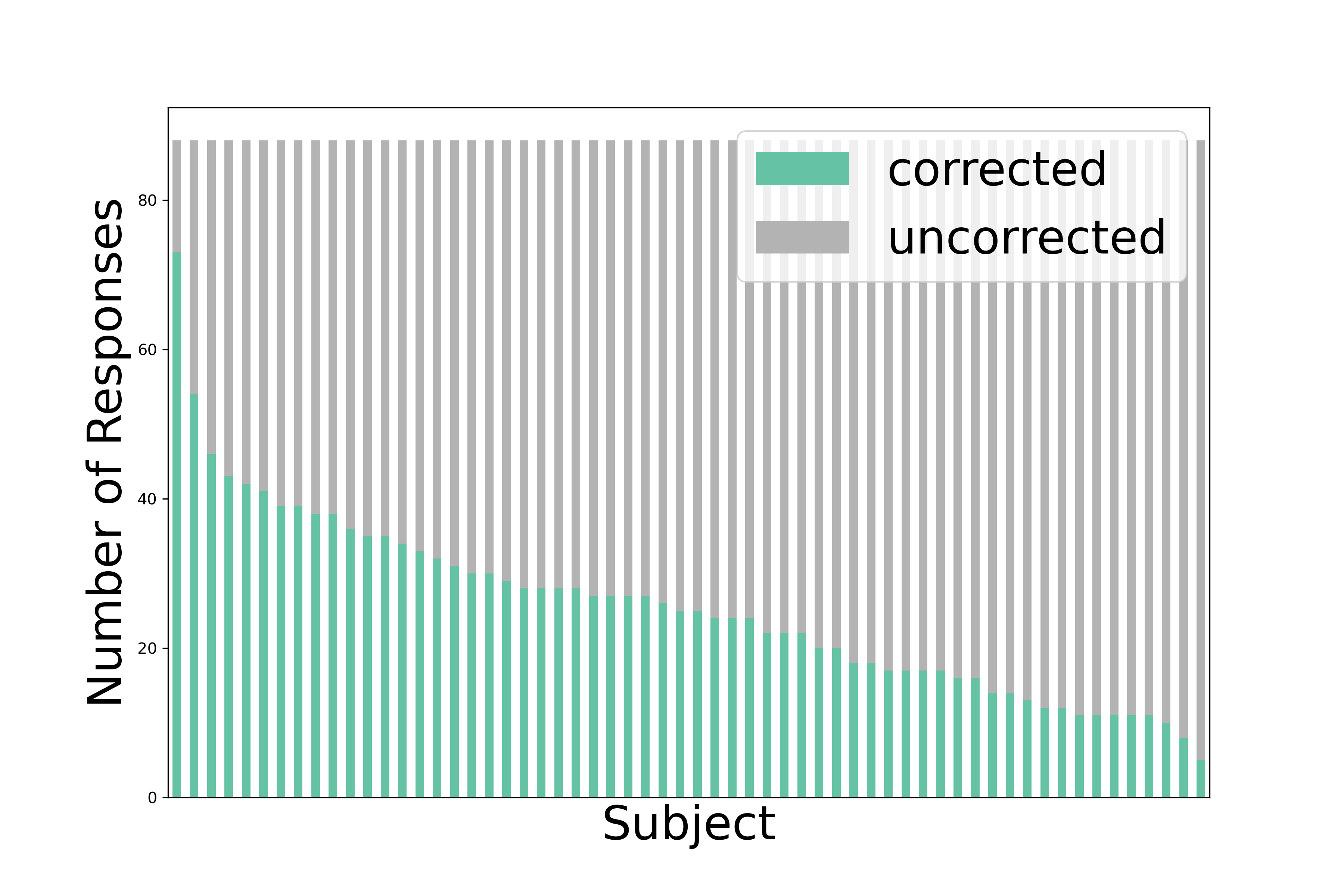}
    \caption{The number of corrected and uncorrected responses for each subject, sorted by the number of corrected responses.}
    \label{fig:response_subject}
\end{figure}

\begin{figure*}[htb!]
    \centering
    \includegraphics[width=0.7\linewidth]{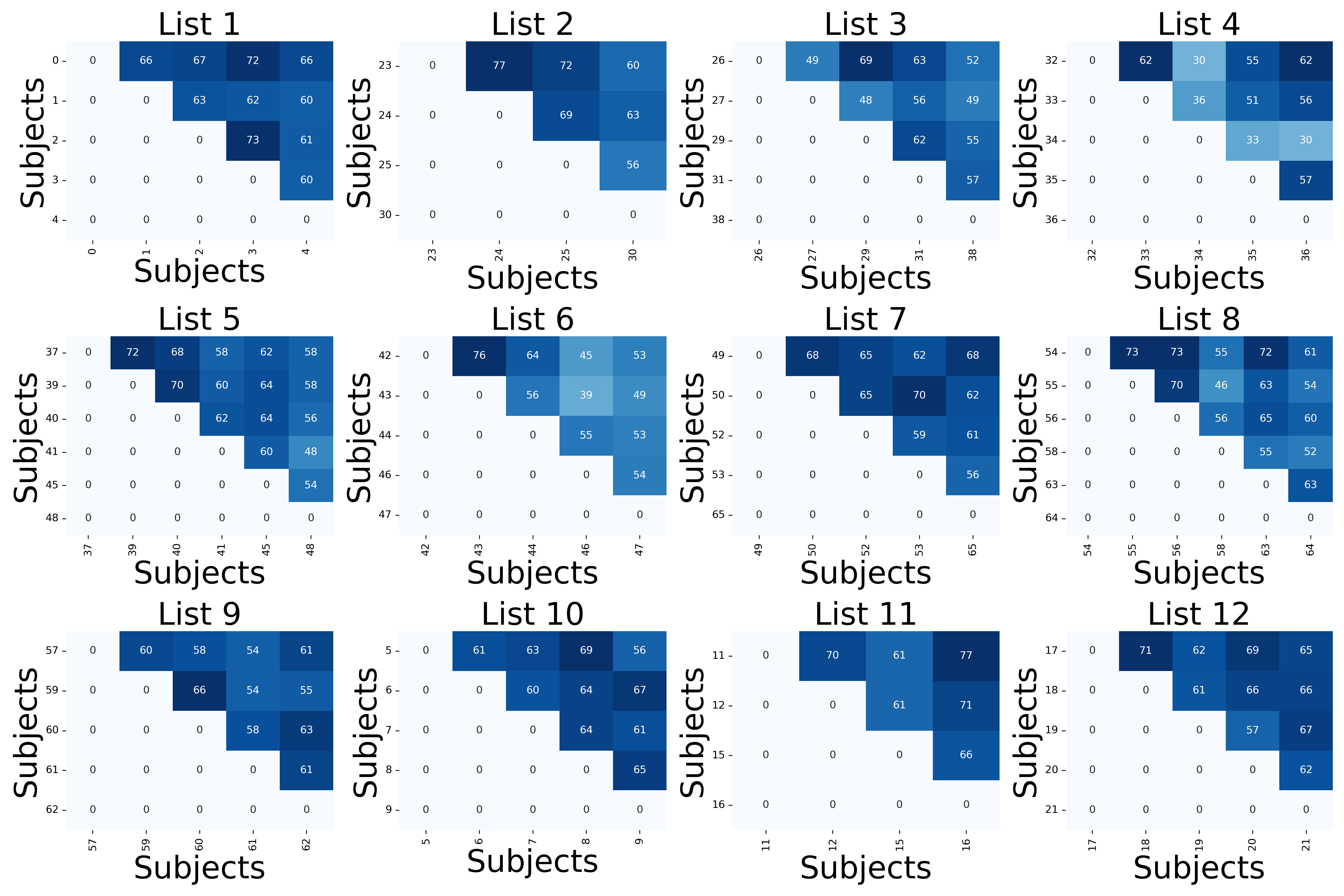}
    \caption{Pairwise agreement scores between subjects across 12 different lists of items.}
    \label{fig:irr}
\end{figure*}

Figure \ref{fig:response_subject} shows the number of corrected and uncorrected responses by each subject. There is variability across subjects, with some participants correcting a substantial proportion of responses while others made few or no corrections. The average Cohen's Kappa across all lists is 0.213, indicating fair agreement between subjects. Figure \ref{fig:irr} displays pairwise agreement scores between subjects across all items (distributed in 12 lists). This suggests that while some systematic agreement is present, variability in responses remains substantial.

\subsection{Computational Metrics}
For words that are corrected by either a speaker or a comprehender and their corresponding corrected counterparts, we calculated the following computational metrics: word predictability, word frequency in SUBTLEXus \citep{brysbaert2009moving}, word length, semantic and phonemic distance between the initial word and the corrected word. We define word predictability as the log probability of the target word given proceeding context, which we estimated from a pre-trained GPT-2 (small, uncased) transformer language model \citep{radford2019language,misra2022minicons}. The semantic distance is calculated using the cosine distance between GloVe word embeddings \citep{pennington2014glove}. To compute the phonemic distance, we first convert the words to their corresponding IPA forms using the Carnegie Mellon Pronouncing Dictionary, and subsequently compute the Levenshtein edit distance between these forms. 

\section{Exploratory Analysis: Modeling Corrections}
The dataset could provide opportunities for the investigation of asymmetric strategies on error correction during language production and comprehension. We present an exploratory analysis that examines the parallels between speaker and comprehender corrections. In particular, we ask (i) whether the lexical properties of the speaker's error and repair are predictive of a comprehender's decision to correct and (ii) whether comprehender's responses over the perceived initial utterance are predictive of the speaker's decision to repair. We model both these choices as functions of the relative \textbf{frequency} and \textbf{predictability} of the initially produced/perceived word and target correction, as well as the \textbf{phonemic} and \textbf{semantic} distance between these forms. 

\subsection{Speaker Model}
A significant challenge in modeling whether or not a speaker has made correction in naturalistic contexts is that instances where a speaker \emph{could} have made a correction but decided not to can only be reliably and accurately determined in highly controlled production studies. Here, we use comprehender corrections to approximate instances where a correction \emph{could} have been made by the speaker. We make this assumption for two reasons. First, comprehender repairs give us a distribution over possible speaker repairs and \emph{may} include the true repair. Furthermore, we assume that the speaker may be incentivized to correct when she believes there is a greater discrepancy between the speaker's intended and comprehender's inferred meaning respectively. 

Consider the following corrections made by the original speaker and three of the comprehenders in the experiment. Based on observed corrections, we define a \colorbox{DeepPlum!50}{\textbf{critical window}} at the position where either a speaker or a comprehender has made a correction.

\begin{examplelist}
\stepcounter{example}
\item \emph{Speaker Corrected}: Well we also in this area seem to have a lot of retirees people who don't want the heat of Florida but don't want the \colorbox{DeepPlum!50}{cold} of the northeast \label{ex:speakerCorrected}
\stepcounter{example}
\item \emph{Comprehender Corrected} \#1: Well we also in this area seem to have a lot of retirees people who don't want the heat of Florida but don't want the \colorbox{DeepPlum!50}{cold} of the northeast \label{ex:compCorrectedMatchSpeaker}
\stepcounter{example}
\item \emph{Comprehender Corrected} \#2: Well we also in this area seem to have a lot of retirees people who don't want the heat of Florida but don't want the \colorbox{DeepPlum!50}{chill} of the northeast \label{ex:compCorrectedNoMatchSpeaker}
\stepcounter{example}
\item \emph{Comprehender Corrected} \#3: Well we also in this area seem to have a lot of \colorbox{DeepPlum!50}{retired} people who don't want the heat of Florida but don't want the heat of the northeast \label{ex:compCorrectedDiffPosition}
\end{examplelist}

For each critical window, we compute the difference in frequency, predictability, phonemic, and semantic representations of the speaker's initial production (e.g., \emph{heat} or \emph{retirees}) and the various comprehender responses in those positions (e.g., \emph{chill} or \emph{cold} in response to \emph{heat} and \emph{retired} in response to \emph{retirees}). We develop the following maximally-converging mixed-effects logistic regression model \citep{barr2013random} to predict correction decisions in speakers:

\noindent {\ttfamily SpeakerCorrected $\sim \Delta$LogProbability + $\Delta$Frequency + Semantic Distance + Phonemic Distance + 1|item + 1|critical 
 window}

where $\Delta$ denotes the \emph{difference} between the frequency and log probability of the speaker's initial production and the comprehender's response. We included random intercepts for item and critical window to account for the participant-level variability at each correction site. We set $\texttt{SpeakerCorrected} = 1$ for critical windows where the speaker did, in fact, make a repair (for example, \ref{ex:speakerCorrected}--\ref{ex:compCorrectedNoMatchSpeaker}). For all other instances, $\texttt{SpeakerCorrected} = 0$ (for example, \ref{ex:compCorrectedDiffPosition}). 

\subsection{Comprehender Model}
When modeling comprehenders' corrections, we restrict the analysis to \textit{speaker corrected} utterances. In particular, we only consider critical windows defined by the speaker's repair (\ref{ex:speakerCorrected}--\ref{ex:compCorrectedNoMatchSpeaker}), thus excluding responses such as (\ref{ex:compCorrectedDiffPosition}).
We compare the word in the critical window in comprehender's final response (\emph{chill} or \emph{cold}) with the word in the perceived initial utterance (\emph{heat}), and annotate it as \textit{corrected} or \textit{uncorrected}. We use the same metrics (relative frequency, predictability, semantic, and phonemic distance), which we calculate over the critical word in the initial utterance (\emph{heat}) and the repair in speaker's final production (\emph{cold}), to predict whether a comprehender will make a correction in their final response.


We then use the following parallel model to predict correction decisions in comprehenders:


\noindent {\ttfamily ComprehenderCorrected $\sim \Delta$LogProbability + $\Delta$Frequency + Semantic Distance + Phonemic Distance + 1|item + 1|critical window + 1|subject}

\begin{figure*}[!htb]
  \centering
    \includegraphics[width=.75\textwidth]{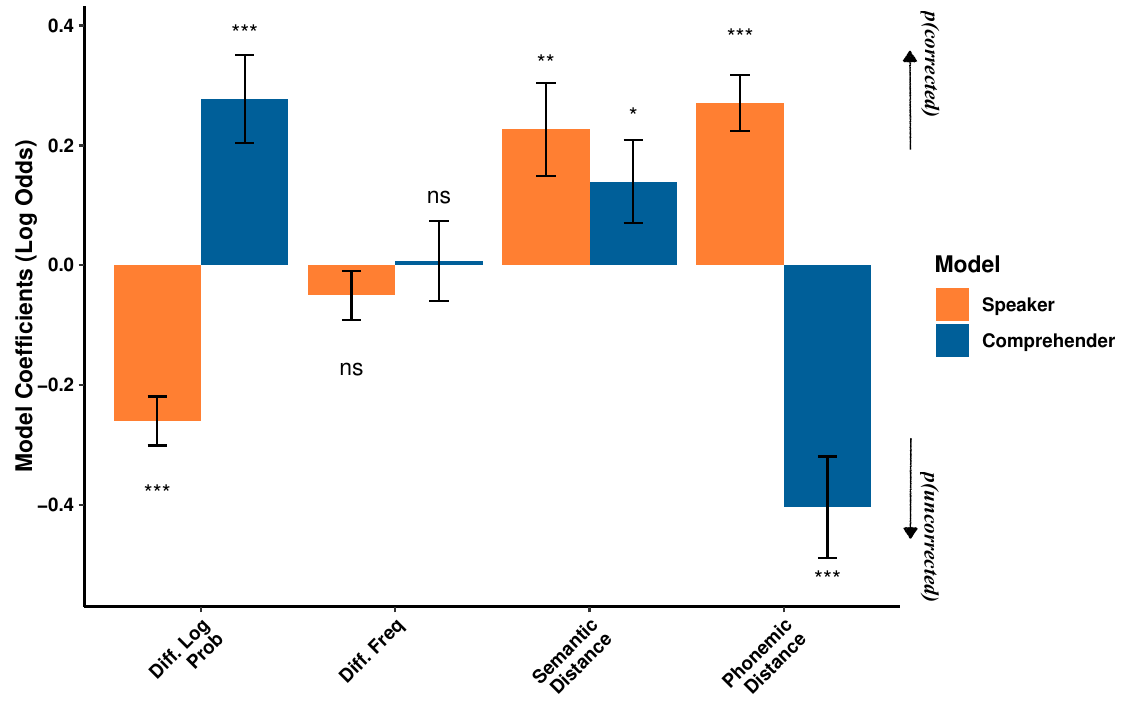}
          \caption{Model coefficients from speaker and comprehender logistic regression models. Error bars denote standard error. Significance: *** ($p < 0.001$), 
          ** ($p < 0.01$), * ($p < 0.05$), \emph{ns} ($p > 0.05$)}
      \label{fig:lexSubsModelResults}
\end{figure*}

\subsection{Results} 
The results from the mixed-effects models reveal key asymmetries in how speakers and comprehenders decide to correct errors, with distinct influences from predictability, frequency, phonemic form, and semantic features. Model coefficients are summarized in Figure~\ref{fig:lexSubsModelResults}.

In the speaker model, the correction behavior is a function of the initially produced/perceived word and the comprehender's final response. When the log probability difference between the comprehender's final inferred response and their initial perceived word is higher, a speaker is less likely to self-correct ($\beta = -0.260, SE = 0.04,  p < 0.01$). This suggests that if comprehenders could recover the intended meaning from the perceived erroneous form, the speaker is less likely to intervene. Additionally, greater semantic distance between the comprehender’s initial perception and their final response increases the likelihood of correction ($\beta = 0.227, SE = 0.08, p < 0.05$), indicating that speakers are more likely to correct when their utterance creates a significant meaning deviation. Phonemic distance exhibits a similar effect, with greater phonemic dissimilarity between the comprehender’s perceived word and the corrected form leading to a higher probability of correction ($\beta = 0.270, SE = 0.05, p < 0.001$). Word frequency does not significantly impact speaker correction behavior ($\beta = -0.05, SE = 0.04, p = 0.21$), suggesting that lexical frequency alone is not a primary driver of corrective processes in production.

In contrast, the comprehenders' corrections are based on the difference between the speaker's erroneous production and their intended repair. Unlike speakers, comprehenders are more likely to correct when the speaker's repair has a much higher log probability than the produced error, as indicated by a significant positive effect of log probability difference ($\beta = 0.278, SE = 0.07, p < 0.01$). This suggests that comprehenders rely more heavily on contextual expectations when detecting errors and overriding them. While semantic distance also increases the likelihood of correction ($\beta = 0.139, SE = 0.07, p < 0.05$), its effect is weaker compared to speakers, implying that comprehenders may is less sensitive to semantic deviations. Crucially, phonemic distance has the opposite effect in comprehension compared to production: while speakers are more likely to correct errors when phonemic distance is large, comprehenders exhibit reduced correction likelihood as phonemic distance increases ($\beta = -0.404, SE = 0.08, p < 0.001$), suggesting the comprehenders are more likely to correct errors that are phonemically similar to the intended meaning. As in the speaker model, frequency does not significantly influence comprehender corrections ($\beta = 0.0068, SE = 0.007, p = 0.92$).

\section{Discussion}
In this study, we develop \emph{SPACER}, a parallel dataset of speech errors and repairs, which is designed to examine how speakers and comprehenders engage in error correction. We identified naturally occurring speech errors and repairs from speech corpus, and used web-based experiments to examine how speech errors are detected and corrected by comprehenders. We conducted exploratory analysis on the asymmetries of error correction strategies between production and comprehension. Specifically, we used linear models to predict whether lexical properties of speaker-produced errors and repairs are predictive of how the errors would be corrected by comprehenders, and whether comprehender's error correction could inform the speaker's self-repair. Our analysis show asymmetries between error correction by speakers and comprehenders. 

Our dataset links production and comprehension and enables a principled comparison between the two modes. Previous studies have largely examined production and comprehension of errors separately \citep{ryskin2018comprehenders,dell1986spreading,levelt1983monitoring,blank2018neural}, making it difficult to assess whether and how these processes might share underlying principles. 
Our dataset bridges the gap by examining how speakers and comprehenders make choices about error correction on the \emph{same} utterance. This parallel structure enables integrated analysis of correction strategies in production and in comprehension, providing a valuable resource for future studies.

Our exploratory analysis reveals key asymmetries in how speakers and comprehenders engage in error correction. Comprehenders are likely to correct errors that are phonemically similar to a more plausible alternative, or when the error is not supported by contextual cues. Speakers, on the other hand, are more likely to self-correct errors that might not be recoverable for comprehenders. The asymmetries might imply potential interplay between comprehension and production, arising from differing demands of the two modalities: while speakers correct their own speech to ensure communicative clarity given the message that they know they want to communicate, comprehenders may be able to use predictability and form-based cues to successfully recover the intended word. The inverse effects of phonemic distance in our models —where comprehenders correct more when errors are phonemically similar to an alternative, while speakers correct more when errors are phonemically distant from the intended meaning -- suggest that interlocutors may engage in complementary error correction strategies. Future work can leverage \textit{SPACER} for more principled computational models that simulate error correction as a rational inference process over various linguistic constraints and cues. 

\section{Limitations}
The dataset is restricted to single-word substitution errors and does not include other common types of speech error such as insertions, deletions, and transpositions. While this allows for a controlled investigation of the error correction process, it may not capture other types of errors and correction strategies in naturalistic communication. Future work could expand SPACER to include a broader range of error types to better understand the full spectrum of production and comprehension repair mechanisms.

A key challenge when studying how speakers may adapt their correction strategies in naturalistic contexts is that the intended lexical target is often difficult to determine. In other words, while speech error corpora provide \emph{positive} examples of speaker corrections, they do not provide instances of \emph{negative} examples i.e., where a speaker \emph{could} have made a repair, but opted against it. Our corpus approach allows us to locate speech errors that have been overtly repaired, but we do not have access to speaker's true communicative goal to identify uncorrected speech errors. We used the distribution of comprehender corrections as an approximation for speaker's communicative goal, to identify errors that could have been corrected by the speaker but were left uncorrected. We acknowledge that the approximation may be different from the speaker's intended meaning.

We used a web-based experiment to investigate how utterances are interpreted and corrected by comprehenders. There are several differences between our experiment and real-time language comprehension. First, comprehenders were presented with the full key sentence, whereas listeners do not have access to the entire utterance at once. Second, comprehenders are given unlimited time to make corrections. In contrast, real-time comprehension is constrained by limited cognitive and processing resources. Third, the text-based presentation method does not provide prosodic and phonetic cues. Comprehenders might rely on disfluencies and pauses to detect errors. Finally, it bears mentioning that the speaker and comprehender corrections are situated in entirely different contexts, separated by both time and space.

\bibliography{imported}

\appendix

\section{Example items and illustration of the pre-processing pipeline} \label{app:preprocess}
\begin{itemize}
\item Original utterance: it depends on whether \textbf{you} whether \underline{we} figure that we have a defense oriented military or an \textbf{aggressive} \underline{aggression} oriented military \label{ex:multipleSpeakerSubs}
\item Frame 1: it depends on whether \textbf{you} whether \underline{we} figure that we have a defense oriented military or an aggression oriented military \label{ex:frame1}
\item Frame 2: it depends on whether we figure that we have a defense oriented military or an \textbf{aggressive} \underline{aggression} oriented military \label{ex:frame2}
\end{itemize}

\section{Instructions and prompts provided to participants in the web-based editing experiment} 
\label{app:experiment}
\begin{figure}[!h]
  \centering
    \includegraphics[width=.5\textwidth]{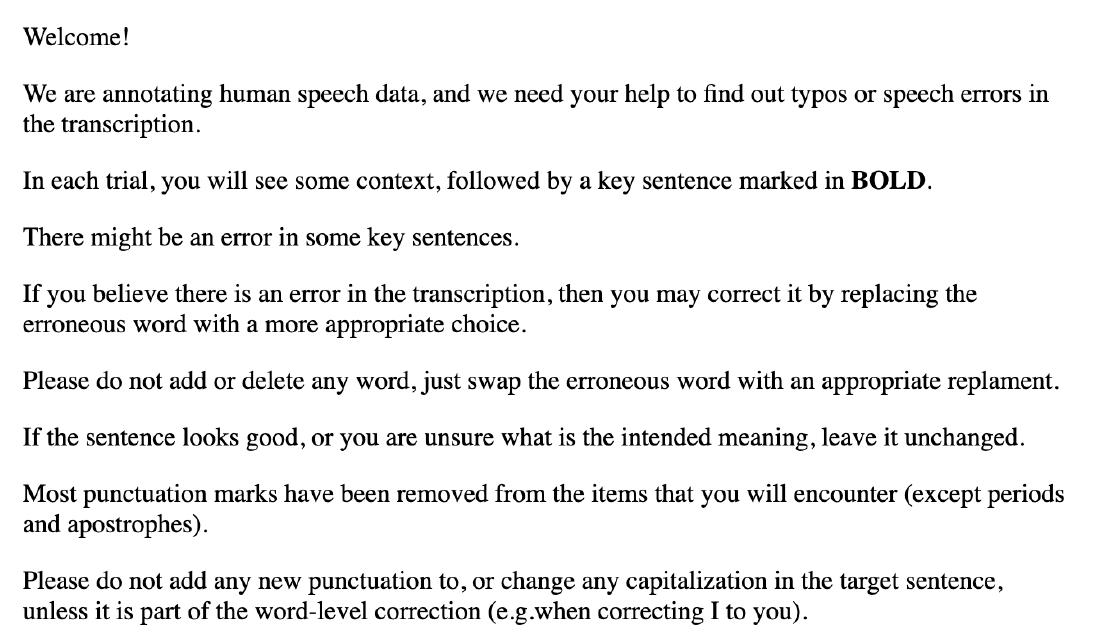}
          \caption{Instructions provided to participants at the beginning of the error correction experiment.}
      \label{fig:Instructions}
\end{figure}

\begin{figure}[!h]
  \centering
    \includegraphics[width=.5\textwidth]{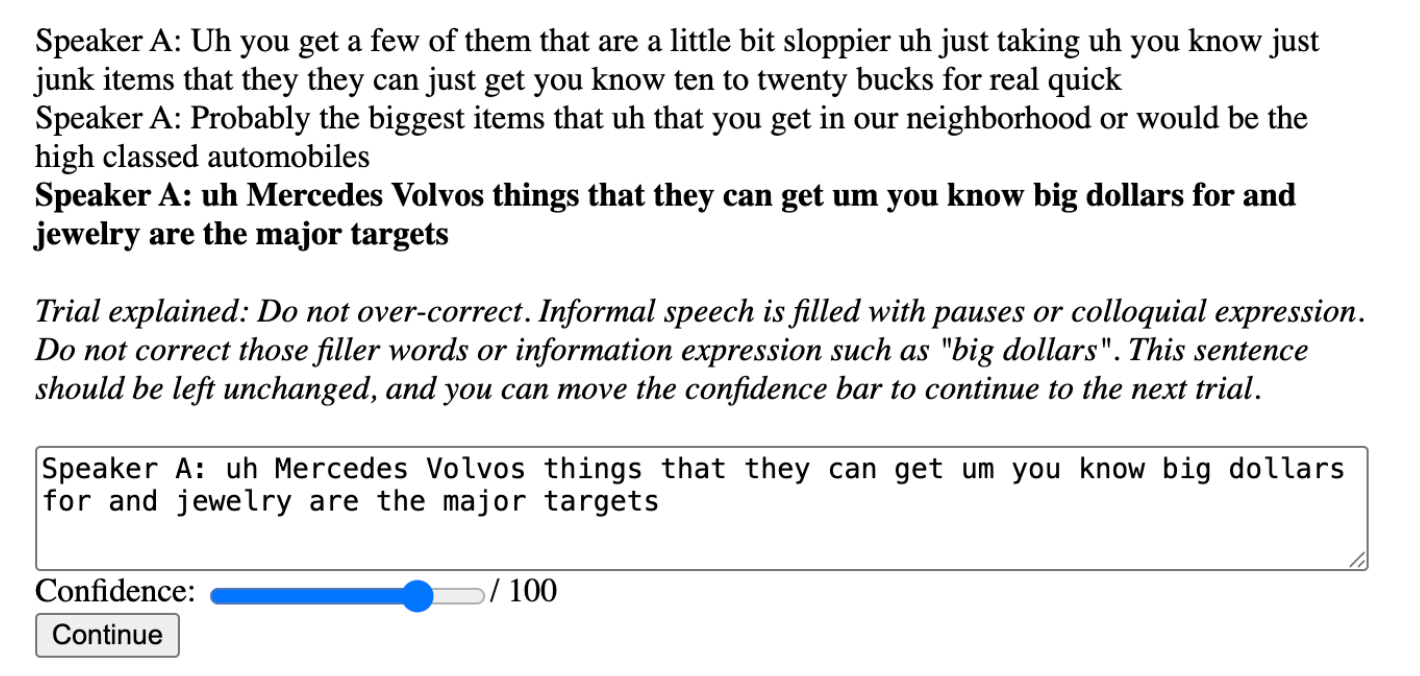}
          \caption{Sample practice trial with feedback to ensure familiarization with the above instructions.}
      \label{fig:PromptToParticipants}
\end{figure}



\end{document}